\documentclass[runningheads]{article}
\usepackage{ijcai25}

\usepackage[
 svgnames,
 table,
]{xcolor}

\usepackage{amssymb}
\usepackage{tikz}
\usepackage{subcaption}
\usepackage{mathtools}
\usetikzlibrary{arrows.meta}
\usetikzlibrary{shapes,snakes}

\usepackage{todonotes}

\newcommand{\R}{\mathbb{R}}

\newcommand{\N}{\mathbb{N}}
\newcommand{\Z}{\mathbb{Z}}
\def\orcidID#1{\smash{\href{http://orcid.org/#1}{\protect\raisebox{-1.25pt}{\protect\includegraphics{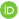}}}}}

\newcommand\Distr{\textit{Distr}}

\newcommand\mdp{\mathcal{M}}
\newcommand\states{S}
\newcommand\Act{A}
\newcommand\Trans{\delta}

\newcommand\dtmc{\mathcal{D}}

\newcommand\mdptup[1] [\refl]{(#1[\states], \allowbreak #1[\Act], \allowbreak #1[\sinit], \allowbreak #1[\Trans])}

\newcommand{\state}{s}
\newcommand{\sinit}{\state_0}
\newcommand\action{a}

\newcommand{\etal}{\emph{et al.}\xspace}

\DeclareMathOperator*{\argmin}{arg\,min}

\newcommand{\abs}[1]{\ensuremath{|#1|}}

\renewcommand{\Pr}{\textit{Pr}}

\usepackage{diagbox}

\DeclareMathSymbol{\shortminus}{\mathbin}{AMSa}{"39}

\usepackage{xspace}

\usepackage{times}
\usepackage{soul}
\usepackage{url}
\usepackage[hidelinks]{hyperref}
\usepackage[utf8]{inputenc}
\usepackage{caption}
\usepackage{graphicx}
\usepackage{amsmath}
\usepackage{amsthm}
\usepackage{booktabs}
\usepackage{algorithm}
\usepackage{algorithmic}

\usepackage{comment}
\usepackage{booktabs}

\usepackage{tikz}
\usepackage[binary-units]{siunitx}

\usepackage{fancyvrb}

\usepackage{wrapfig}

\newtheorem{example}{Example}
\newtheorem{definition}{Definition}
\newtheorem{theorem}{Theorem}
\newtheorem{lemma}{Lemma}
\newenvironment{proofsketch}{%
  \proof}{\endproof}

\usepackage[switch]{lineno}

\title{Counterfactual Strategies for Markov Decision Processes}

\iftrue
\author{
Paul Kobialka$^1$\and
Lina Gerlach$^2$\and
Francesco Leofante$^3$\and
Erika Ábrahám$^2$\and\\
Silvia Lizeth Tapia Tarifa$^1$\And
Einar Broch Johnsen$^1$\\
\affiliations
$^1$University of Oslo, Oslo, Norway\\
$^2$RWTH Aachen University, Germany\\
$^3$Imperial College London, United Kingdom\\
\emails
\{paulkob, sltarifa, einarj\}@ifi.uio.no,
f.leofante@imperial.ac.uk,\\
\{gerlach, abraham\}@cs.rwth-aachen.de
}
\fi

\begin{document}

\maketitle

\begin{abstract} 
  Counterfactuals are widely used in AI to explain how minimal changes
  to a model's input can lead to a different output.  However,
  established methods for computing counterfactuals typically focus on one-step
  decision-making, and are not directly applicable to sequential
  decision-making tasks.  This paper fills this gap by introducing
  counterfactual strategies for Markov Decision Processes (MDPs).
  During MDP
  execution, a strategy decides which of the enabled actions (with
  known probabilistic effects) to execute next. Given an initial
  strategy that reaches an undesired outcome with a probability above
  some limit, we identify minimal changes to the initial strategy to
  reduce that probability below the limit. We encode such
  counterfactual strategies as solutions to non-linear optimization
  problems, and further extend our encoding to synthesize diverse
  counterfactual strategies. We evaluate our approach on four
  real-world datasets and demonstrate its practical viability in sophisticated sequential decision-making
  tasks.
\end{abstract}

\section{Introduction}
\label{sec:introduction}






Consider an application procedure in which clients who want to obtain
a loan, interact with a bank to establish their eligibility.  Although
established prediction
methods~\cite{leo2019machine,teinemaa2019outcome} can be used to
filter for eligible clients, the overall application procedure is far
from automated.  In practice, to receive a loan from the bank, a
client must follow a complicated application procedure, involving,
e.g., multiple consultations with loan advisors, providing various
documents and filling out complex forms.  Eligible but impatient
clients are prone to abandoning the application procedure before they
receive their loan, causing losses for both parties: the client spent
time without reaching their goal and the bank invested resources
without return. Markov Decision Processes (MDPs)~\cite{baier2008principles}
can be used to model such procedures and improve their transparency;
however, methods are currently lacking to address questions about
recourse and process improvement; e.g., \emph{what would enable an
  ineligible client to obtain a loan?} and \emph{how can we simplify
  the application procedure for eligible clients?}\looseness=-1

\emph{Counterfactual explanations} can help answer such questions by showing how minimal changes in user applications would lead to a desired change of the output, e.g., to make the client
eligible for the previously refused loan.  
However,
most available methods for computing counterfactuals target one-step
prediction tasks~\cite{guidotti2024counterfactual}, and are not
applicable to sequential decision making settings, where the output of a process is determined by a sequence of steps.


\textbf{Contributions.} To fill this gap, we propose a method to compute counterfactual explanations for sequential
decision making processes modeled by an MDP. In each state of these non-deterministic discrete-time
models, a finite number of actions with a known probabilistic effect
are enabled. During execution, strategies decide which enabled
action to execute next. Given an initial strategy that visits some
undesired states with a probability above some limit, we propose to compute explanations in terms of \emph{counterfactual strategies} that reduce the
reachability probability below this limit, while staying as close to
the initial strategy as possible.  To formalize these requirements, we
introduce a \emph{distance measure} $d$ on strategies and encode
counterfactual strategy synthesis as nonlinear optimization
problems.  By enforcing that only user-controllable actions can be
selected in the MDP, we ensure that the resulting counterfactual
strategies are indeed \emph{actionable}. Furthermore, we extend our encoding to compute not only a single solution but a
collection of counterfactual strategies, which are optimized for
\emph{diversity}, as several studies emphasize that providing
different counterfactuals is key for user
understanding~\cite{russell2019efficient,mothilal2020explaining,bove2023investigating}. The method is evaluated on four real-world datasets. The
evaluation shows that the method is computationally feasible,
and can synthesize counterfactual strategies for sophisticated
sequential decision making tasks, modeled as MDPs with 
thousands of states and ten thousands of transitions.

In summary, our main contributions are (1) to introduce counterfactual
strategies for MDPs as post-hoc explanations for sequential decision
making, (2) to encode a counterfactual strategy synthesis problem for
MDPs as a nonlinear optimization problem, (3) to synthesize diverse
counterfactual strategies, and (4) to experimentally evaluate the
feasibility and performance of our counterfactual strategy synthesis
method.

\paragraph{Outline.} After recalling some background on MDPs,
nonlinear optimization and counterfactuals in
Section~\ref{sec:prelim}, we formalize counterfactual strategies for
MDPs in Section~\ref{sec:ce_mdp}, and present our encoding of the MDP
counterfactual synthesis problem as a non-linear optimization problem
in Section~\ref{sec:computing_ce}. We evaluate our approach in
Section~\ref{sec:experiments}, discuss related work in
Section~\ref{sec:related_work}, and conclude the paper in
Section~\ref{sec:conclusion}.


\section{Preliminaries}
\label{sec:prelim}
A (discrete probability) \emph{distribution} is a function $\mu \colon X \rightarrow [0,1]$ with a discrete domain $X$ such that
$\sum_{x \in X} \mu(x) = 1$.
We write $\Distr(X)$ for the set of distributions with domain $X$.
Given two distributions $\mu_1$ and $\mu_2$ with domain $X$, their \emph{total variation distance} is $\Delta(\mu_1,\mu_2) = \frac{1}{2} \sum_{x \in X} \abs{ \mu_1(x) - \mu_2(x)}$~\cite[Prop. 4.2]{levin2017markov}, where $\abs{\cdot}$ stays for the absolute value.

For $n\in\N$,  $v \in \R^n$ and $i =1,\ldots, n$, we denote the $i$th element of $v$ as $v_i$,
and use the standard norms 
$\|v\|_{0} = | \{v_i \mid v_i \neq 0 \}|$, 
$\|v\|_{1} = \sum_i \abs{v_i}$,
and
$\|v\|_{\infty} = \max_i  \abs{v_i} $.

\subsection{Markov Decision Processes}
A \emph{Markov decision process} (MDP) $\mdp$ is a tuple
$\langle \states, \Act, \sinit, \Trans \rangle$, where $\states$ is a
finite set of states, $\Act$ is a finite set of actions,
$\sinit\in\states$, and
$\Trans \colon \states \times \Act \rightharpoonup \Distr(\states)$ is
a partial function. For each state $s\in\states$, let $\Act(s)$ be the
set of all actions $a\in\Act$ for which $\Trans(s,a)$ is defined; we
require $\Act(s)\not=\emptyset$ and say that the actions in $\Act(s)$
are \emph{enabled} in $s$. By $\Trans(s,a,s')$ we denote
$\Trans(s,a)(s')$ if $\Trans(s,a)$ is defined and $0$ otherwise.

A (finite or infinite) \emph{path} $\tau$ of $\mdp$ is a non-empty sequence of alternating states and actions $s_0 a_0 s_1\dots $ such that $\Trans(s_j,a_j,s_{j+1})>0$ for all $j\geq 0$.
The \emph{cylinder set} $\textit{Cyl}(\hat{\tau})$ of a finite path $\hat{\tau}$ is the set of all infinite paths with $\hat{\tau}$ as a prefix.
Let $\Omega_{\mdp}(s)$ and and $\Omega^{\textit{fin}}_{\mdp}(s)$ be the set of all infinite respectively finite paths of $\mdp$ starting in the state $s \in \states$. 

A state $t \in \states$ can be \emph{reached} from state $s \in \states$ if there exists a finite path from $s$ to $t$.
We use $\textit{Reach}(t)$ to denote the set of all states from which $t$ can be reached.

A (memoryless) \emph{strategy} is a function
$\sigma \colon \states \to \Distr(\Act)$ that maps states to distributions over
actions with $\sigma(\state)(\action) = 0$ for all $\state \in \states$ and $\action \in\Act{\setminus} \Act(\state)$.
We denote the set of strategies for $\mdp$ by $\Sigma_\mdp$; we omit the index if it is clear from the context.
Given two strategies $\sigma, \sigma' \in \Sigma$, we overload notation and define their \emph{distance vector}
as 
$\Delta(\sigma,\sigma') = (\Delta(\sigma(s), \sigma'(s)))_{s \in \states}$ (entries in fixed but arbitrary order).

Applying a strategy $\sigma$ to an MDP $\mdp$ induces a deterministic model. 
Thus we omit the actions and define the \emph{discrete-time Markov chain (DTMC) induced by $\sigma$ on $\mdp$} as the tuple $\mdp^\sigma =\langle \states, \sinit, \Trans^{\sigma} \rangle$ with $\states$ and $\sinit$ as before and $\Trans^{\sigma} \colon \states \times \states \to [0,1]$ with $\Trans^\sigma(s, s') = \sum_{a \in \Act} \sigma(s)(a)\cdot\Trans(s, a, s')$ for all $s,s' \in \states$.\looseness=-1

We associate with $\dtmc {=} \mdp^\sigma$ the \emph{probability space}
$(\Omega_\dtmc(\sinit), \allowbreak \{\bigcup_{\hat{\tau} \in R} \textit{Cyl}(\hat{\tau}) \mid R~\subseteq~\Omega^{\textit{fin}}_{\dtmc}(\sinit)\}, \allowbreak \Pr_\dtmc(\sinit))$
where the probability of the cylinder set of a finite path $\hat{\tau} = s_0\ldots s_n$ is $\Pr_\dtmc(\sinit)(\textit{Cyl}(\hat{\tau})) = \prod_{i=1}^{n} \Trans(s_{i-1}, s_i)$.
%
By $\Pr_{\dtmc}(s_0, t)$ we denote the probability of reaching state $t \in \states$ from $s_0$ in $\dtmc$.

\begin{figure}[t!]
    \centering
        \begin{subfigure}[b]{\linewidth}
            \centering
            \resizebox{0.9\linewidth}{!}{
            \tikzstyle{line} =[line width=2\unitlength, black!60, >=latex]
\begin{tikzpicture}[
  rflow/.style = {thick,->},
  circstyle/.style = {shape=circle,draw=black, minimum size=5pt, inner sep=0pt,thick},
  eclipstyle/.style = {shape=ellipse,draw=black,thick},
  ]
    \node[eclipstyle] (s0) at (0,0) {$s_0$};
    \node[circstyle] (s0consult) at (2.4,-0.8) {};
    \node[circstyle] (s0apply) at (-1.1,-0.8) {};
    
    \node[eclipstyle] (Application) at (-3,-2) {Application};
    \node[circstyle] (applicationProvider) at (-3,-5) {};

    \node[eclipstyle] (Error) at (1,-1.5) {Error};
    \node[circstyle] (errorConsult) at (0.52,-2.2) {};
    \node[circstyle] (errorQuit) at (3.0,-2.2) {};

    \node[eclipstyle] (Consultation) at (0,-3) {Consultation};
    \node[circstyle] (consultationApply) at (0,-3.8) {};
    \node[circstyle] (consultationQuit) at (2.3,-3.8) {};

    \node[eclipstyle] (Application+) at (0,-4.8) {$\text{Application}^+$};
    \node[circstyle] (application+Provider) at (0,-5.8) {};

    \node[eclipstyle] (Rework) at (0,-6.6) {Rework};
    \node[circstyle] (reworkSubmit) at (0,-7.5) {};
    \node[circstyle] (reworkQuit) at (2,-7.5) {};

    \node[eclipstyle] (Resubmit) at (0,-8.4) {Resubmit};
    \node[circstyle] (resubmitProvider) at (0,-9.2) {};

    \node[eclipstyle] (Granted) at (-3,-10) {Granted} ;
    \node[eclipstyle] (Rejected) at (3,-10) {Rejected} ;

    \path [rflow] (s0) edge node[left, near start, xshift=-1mm,yshift=1mm] {Apply} (s0apply);
    \path [rflow] (s0) edge node[right, near start, xshift=3mm, yshift = 1mm] {Consult} (s0consult);
    \path [rflow] (s0apply) edge node[left, near start, xshift=-3mm] {$0.95$} (Application);
    \path [rflow] (s0apply) edge node[right, near start, xshift=3mm, yshift=1mm] {$0.05$} (Error);
    \path [rflow] (s0consult) edge[bend left = 35] node[near end,xshift=2mm,yshift=-1mm] {$1$} (Consultation);

    \path [rflow] (Application) edge node[left] {Provider} (applicationProvider);
    \path [rflow] (applicationProvider) edge[bend right = 40] node[left] {0.5} (Granted);
    \path [rflow] (applicationProvider) edge[bend right = 20, near start] node[right, xshift=1mm] {0.5} (Rework);

    \path [rflow] (Error) edge node[left, xshift=-2mm] {Consult} (errorConsult);
    \path [rflow] (Error) edge node[right, xshift=1mm, yshift=2mm] {Quit} (errorQuit);
    \path [rflow] (errorConsult) edge node[left, xshift=-1mm,yshift=1mm] {$1$} (Consultation);
    \path [rflow] (errorQuit) edge[bend left = 55] node[right] {$1$} (Rejected);

    \path [rflow] (Application+) edge node[right, xshift=1mm,yshift=-1mm] {Provider} (application+Provider);
    \path [rflow] (application+Provider) edge node[right, xshift=1mm,yshift=1mm] {0.1} (Rework);
    \path [rflow] (application+Provider) edge[bend right = 45] node[left,xshift=-1mm] {0.9} (Granted);

    \path [rflow] (Rework) edge node[left, near start,xshift=-1mm,yshift=-1mm] {Submit} (reworkSubmit);
    \path [rflow] (Rework) edge node[right, near start, xshift=1mm, yshift=1mm] {Quit} (reworkQuit);
    \path [rflow] (reworkSubmit) edge node[right, xshift=1mm,yshift=1mm] {1} (Resubmit);
    \path [rflow] (reworkQuit) edge[bend left = 25] node[right] {1} (Rejected);

    \path [rflow] (Consultation) edge node[right, near start, xshift=3mm,yshift=1mm] {Quit} (consultationQuit);
    \path [rflow] (Consultation) edge node[right, xshift=1mm,yshift=-1mm] {Apply} (consultationApply);
    \path [rflow] (consultationQuit) edge[bend left = 35] node[right] {1} (Rejected);
    \path [rflow] (consultationApply) edge node[right] {1} (Application+);

    \path [rflow] (Resubmit) edge node[left, xshift=-1mm,yshift=-1mm] {Provider} (resubmitProvider);
    \path [rflow] (resubmitProvider) edge node[right, near end, xshift=-2mm,yshift=3mm] {0.2} (Rejected);
    \path [rflow] (resubmitProvider) edge node[left, near end, xshift=2mm,yshift=3mm] {0.8} (Granted);

    \path [rflow] (Granted) edge[loop left] node[left] {1} (Granted);
    \path [rflow] (Rejected) edge[loop right] node[right] {1} (Rejected);
\end{tikzpicture}

            }
        \caption{MDP model. The only service provider action \textit{Provider} appears deterministically, thus the user has full strategy control.}
    \label{fig:loan-example}
    \end{subfigure}%
    
    \begin{minipage}[b]{\linewidth}
    \centering
    \begin{subfigure}[b]{\linewidth}
\smallskip
      \begin{tabular}{|c|c|c|c|c|c|}\hline
\diagbox{state $s$}{action $a$} & Apply & Consult & Quit & Submit \\\hline
        $s_0$ & 1 & 0 & 0 & 0\\
        Error & 0 & 0.2 & 0.8 & 0 \\
        Consultation & 0 & 0 & 1 & 0 \\
        Rework & 0 & 0 & 0.7 & 0.3 \\\hline
      \end{tabular}

      \caption{Probability values $\sigma(s)(a)$ of the impatient client strategy $\sigma$.}
    \label{fig:ce_initial}
    \end{subfigure} \hfill
    \begin{subfigure}[b]{\linewidth}
\smallskip
      \begin{tabular}{|c|c|c|c|c|c|}\hline
\diagbox{state $s$}{action $a$} & Apply & Consult & Quit & Submit \\\hline
        $s_0$ & 1 & 0 & 0 & 0\\
        Error & 0 & 0.2 & 0.8 & 0 \\
        Consultation & 0 & 0 & 1 & 0 \\
        Rework & 0 & 0 & \bf 0.14 & \bf 0.86 \\\hline
      \end{tabular}

      \caption{Counterfactual strategy $\sigma^*$ for the impatient client.}
        \label{fig:ce_impatient}
        \end{subfigure}
    \end{minipage}%
    \caption{Running example of a loan application procedure.}
\end{figure}

\begin{example}
    \label{ex:mdp_def}
    Figure~\ref{fig:loan-example} shows an MDP model $\mdp$ of a loan
    application procedure.  Starting in $s_0$, the client either
    directly fills out an application or requests a
    consultation to increase the probability of direct acceptance.
    However, when independently filling out the application, there is
    a $5\%$ chance to make a mistake in the application, which requires a
    consultation to fix.
    If the application is not accepted directly,  it can be reworked before it is  evaluated.
    The client may decide to quit the application procedure after making a mistake in the form, after the consultation, or if the application is not directly accepted.
    The behavior of the service provider  is captured by several occurrences of the \textit{Provider} action.

    The client's goal is to receive a loan, i.e.\ reach the
    \textit{Rejected} state with a probability of at most $20\%$.
    For an impatient client who directly fills out the
    application using strategy $\sigma \in \Sigma_\mdp$ in
    Fig.~\ref{fig:ce_initial}, the probability of reaching
    \textit{Rejected} is
    $Pr_{\mdp^\sigma}(s_0, \textit{Rejected}) = 0.411$.
\end{example}

\subsection{Non-linear Optimization}
\emph{Mixed Integer Quadratically Constrained Quadratic Problems (MIQCQPs)}~\cite{billionnet2016exact} are a class of non-linear optimization problems, where the objective function and the constraints are at most quadratic in variables with real and integer domains.
Formally, an MIQCQP has the form
\begin{equation*}
    \begin{array}{lll}
        \min \quad & f_0(x) \\
        \text{subject to } & f_i(x) \leq b_i & \textit{for\ \ } i = 1, \dots, m,\\
        & 0 \leq x_j \leq u_j &  \textit{for\ \ } x_j \in V_{\Z} \cup V_{\R}, \\
        & x_j \in \Z &  \textit{for\ \ } x_j \in V_{\Z}, \\
        & x_j \in \R &  \textit{for\ \ } x_j \in V_{\R},
    \end{array}
\end{equation*}
where $m\in\N$ is the number of constraints, $x=(x_1,\ldots,x_n)$ are the variables divided into the sets $V_{\Z}$ and $V_{\R}$ of integer- respectively real-valued variables,
$f_i(x) = x^T Q_i x + c^T_i x$ for all $ i \in \{0, \dots, m\}$ with symmetric matrices $Q_i \in \R^{n \times n}$ and $c_i \in \R^n$.
Bounds conform to the variable domains, i.e. $u_j \in \Z$ for $x_j \in V_{\Z}$, and $u_j \in \R$ for $x_j \in V_{\R}$. 
%
%
MIQCQPs are not convex and in general hard to solve~\cite{billionnet2016exact,garey1979computers}. 

\subsection{Counterfactual Explanations}
Informally, counterfactual explanations answer the question ``\emph{If $A$ were true, would $C$ have been true?}'' by providing a \emph{counterfactual antecedent} $A$ such that under its observation the \emph{counterfactual consequent} $C$ would have evaluated to true~\cite{balke1994counterfactual}. Our notion of counterfactual strategies echoes common formalizations in machine learning~\cite{russell2019efficient,mothilal2020explaining,guidotti2024counterfactual,molnar2020interpretable}, which typically define counterfactual explanations as follows. 
For a set of classes $C$, a classifier $f:\R^n\rightarrow C$, and an input $x\in\R^n$, a counterfactual is a closest input to $x$ w.r.t.\ a distance measure $d$ that yields a desired class $c \in C$:
\[
\argmin_{x'} d(x,x') \quad \text{subject to} \quad
f(x') = c.
\]


This basic formulation of counterfactuals requires $x'$ to be close to the initial input $x$,
to ensure that the changes suggested by the counterfactual are realistic.\footnote{Note that if $f(x)=c$ then $x'=x$ is a counterfactual.} Further properties might be required for counterfactual explanations (see, e.g., the recent survey~\cite{guidotti2024counterfactual}). In the next section we discuss some of them, and map them to the MDP setting.




\section{Counterfactual Explanations for MDPs}
\label{sec:ce_mdp}

In this section, we first discuss desired properties of counterfactuals (as formalised in the machine learning literature) and how they translate to MDPs. Based on these, we then introduce our definition of counterfactual strategies for MDPs.
We refine our definition to account for different notions of distances between the counterfactual strategy and the initial strategy, while also extending the running example.

We consider the following four desired properties of counterfactual explanations from machine learning (ML)~\cite{gajcin2024redefining,guidotti2024counterfactual} and translate them to MDPs:
\begin{itemize}
\item Validity: \emph{ML:} The counterfactual does change the classification to the desired class, i.e. $f(x') = c$. \emph{MDP:} Following the counterfactual strategy reduces the probability of reaching $t$ below a given threshold.
\item Proximity: \emph{ML:} The distance between initial input and counterfactual is minimal.
  \emph{MDP:} The distance between initial strategy and counterfactual strategy is minimal.
    \item Actionability: \emph{ML:} Only features from a set of actionable features are mutated.
    \emph{MDP:} Only actions controlled by the user are altered in the counterfactual strategy.
    \item Sparsity: \emph{ML:} The number of changed features is minimal. \emph{MDP:} The number of actions changed in the counterfactual strategy is minimal.
\end{itemize}


To define distance measures for strategies, 
for any two strategies $\sigma$, $\sigma' \in \Sigma$, let
\begin{align*}
    d_{0}(\sigma, \sigma') &:= \|\Delta(\sigma, \sigma')\|_0 \\
    d_{1}(\sigma, \sigma') &:= \|\Delta(\sigma, \sigma')\|_1 \ / \ |S| \\
    d_{\infty}(\sigma, \sigma') &:= \|\Delta(\sigma, \sigma')\|_\infty 
\end{align*}
\noindent
where $|S|$ is the number of states. Here, $d_{0}$ captures the sparsity of the counterfactual by measuring the number of states where a decision was changed, while $d_{1}$ and $d_{\infty}$  address proximity by measuring the average, respectively maximal, changes over all states between counterfactual and input strategy.

A \emph{strategy distance measure} for $\mdp$ is a function $d:\Sigma_{\mdp}\times \Sigma_{\mdp}\rightarrow \R$ using some $ r_0, r_1, r_\infty\in \R$ to define
$
d(\sigma, \sigma')=
r_0 \cdot d_{0}(\sigma, \sigma') +
r_1 \cdot d_{1}(\sigma, \sigma') +
r_\infty \cdot d_{\infty}(\sigma, \sigma') 
$  for $\sigma,\sigma'\in\Sigma_\mdp$.

Now we are ready to define counterfactual strategies for MDPs.

\begin{definition}[Counterfactual Strategy] 
\label{def:cs}
Assume an MDP $\mdp = \mdptup$, a strategy $\sigma \in \Sigma_{\mdp}$, a bound $\gamma \in [0,1]$, and a target state $t \in S$ such that $Pr_{\mdp^\sigma}(\sinit, t) > \gamma$.
Let furthermore $d$ be a strategy distance measure for $\mdp$. 
We call a strategy $\sigma^* \in \Sigma_{\mdp}$
a \emph{counterfactual strategy} to $\sigma$ (under $d$ for reaching $t$ within $\gamma$ in $\mdp$) if (i) $Pr_{\mdp^{\sigma^*}}(\sinit, t) \leq \gamma$ and (ii) $d(\sigma, \sigma^*) \leq d(\sigma, \sigma')$ for all $\sigma' \in \Sigma_\mdp$ with $\Pr_{\mdp^{\sigma'}}(s_0, t) \leq \gamma$.
\end{definition}

\begin{example}
    \label{ex:mdp_ce}
    Consider again the MDP $\mdp$ and the strategy $\sigma$ from Ex.~\ref{ex:mdp_def} with strategy distance measure $d(\sigma', \sigma'') \coloneqq d_{0}(\sigma', \sigma'') + d_{1}(\sigma', \sigma'') + d_{\infty}(\sigma', \sigma'')$.
    Strategy $\sigma^* \in \Sigma_\mdp$ from Fig.~\ref{fig:ce_impatient} is a counterfactual strategy to $\sigma$ under $d$ for reaching $\textit{Rejected}$ within $\gamma = 0.2$ in $\mdp$ by asking the client to continue after \textit{Rework}.
    
    
\end{example}

To reduce computational cost, we also define $\epsilon$-counterfactual strategies by replacing the requirement of smallest distance by the requirement of bounded distance.

\begin{definition}[$\epsilon$-Counterfactual Strategy]
\label{def:valid_cs}
Assume an MDP $\mdp = \mdptup$, a strategy $\sigma \in \Sigma_{\mdp}$, a bound $\gamma \in [0,1]$, and a target state $t \in S$ such that $Pr_{\mdp^\sigma}(\sinit, t) > \gamma$.
Let furthermore $d$ be a strategy distance measure for $\mdp$  and $\epsilon > 0$. 
We call a strategy $\sigma^* \in \Sigma_{\mdp}$
an \emph{$\epsilon$-counterfactual strategy} to $\sigma$ (under $d$ for reaching $t$ within $\gamma$ in $\mdp$) if (i) $Pr_{\mdp^{\sigma^*}}(\sinit, t) \leq \gamma$ and (ii) $d(\sigma, \sigma^*) \leq \epsilon$.
\end{definition}


In this paper we focus on counterfactual strategies, but our methods can be easily adapted to $\epsilon$-counterfactual strategies, which need less computational effort because the optimality criterion is dropped.

\begin{example}
    Strategy $\sigma^*$ from Ex.~\ref{ex:mdp_ce} changes the decision only in state \textit{Rework}, thus $\Delta(\sigma, \sigma^*) = (0,0,0,0.56)$.
    Therefore, strategy $\sigma^*$ is a $0.56$-counterfactual strategy under $d_{\infty}$, $0.14$-counterfactual strategy under  $d_{1}$ and a $1$-counterfactual strategy under $d_0$.
    Only all three measures combined reveal an accurate picture of the changes required for adapting to the counterfactual.
    Note that for $\epsilon < 0.51$, there exists no valid counterfactual strategy under $d_{\infty}$ for $\gamma = 0.2$.
\end{example}

\section{Computing Counterfactual Strategies}
\label{sec:computing_ce}
We propose to generate counterfactual strategies $\sigma^*$ by solving \emph{non-linear optimization problems}, minimizing the distance $d(\sigma,\sigma')=r_0\cdot d_0(\sigma,\sigma')+r_1\cdot d_1(\sigma,\sigma')+r_\infty \cdot d_\infty(\sigma,\sigma')$ to the initial strategy $\sigma$ over all strategies $\sigma'$ that reach the target state $t$ with a probability below the given limit of $\gamma$:
\[
\argmin_{\sigma' \in \Sigma_{\mdp}} d(\sigma,\sigma') \quad \text{subject to} \quad Pr_{\mdp^{\sigma'}}(s_0, t) \leq \gamma.
\]
To formalize the above optimization problem in arithmetic terms, we use for each $s \in S$ and $a\in\Act(s)$ the following \emph{real} variables: (1)  $p_{sa}$ to encode the probability $\sigma'(s)(a)$ that in the state $s$ the counterfactual strategy $\sigma'$ chooses the action $a$; (2) $p_s$ to encode the probability of reaching $t$ from $s$ in $\mdp^{\sigma'}$; (3) $\Delta_s$ to encode $\Delta(\sigma(s),\sigma'(s))$; (4) $D_{\bowtie}$ for $\bowtie \in \{0,1,\infty\}$ to encode the distances $d_{\bowtie}(\sigma,\sigma')$.
In addition, we introduce for each state $s$ an \emph{integer} variable $i_s \in \{0, 1\}$, whose value is $1$ iff $\sigma$ and $\sigma'$ define different distributions at state $\state$.
For fixed input MDP $\mdp=\langle \states, \Act, \sinit, \Trans \rangle$, state $t$, limit $\gamma$, strategy $\sigma$, and real coefficients $r_0$, $r_1$ and $r_\infty$, the encoding is as follows: 
%
\begin{eqnarray}
    \min \quad r_0 \cdot D_0 + r_1 \cdot D_1 + r_\infty \cdot D_\infty\label{eq:min}
\end{eqnarray}
\quad subject to
\begin{eqnarray}
    \forall s \in S, a \in \Act(s): && 0  \leq p_{sa} \leq 1 \label{eq:psa}\\
    \forall s \in S: && \sum_{a\in\Act(s)} p_{sa}  = 1 \label{eq:sum}\\
    && p_t  = 1 \label{eq:pt} \\
    \forall s \in S {\setminus} \textit{Reach}(t): && p_s  = 0 \label{eq:pinit}\\
    \forall s \in \textit{Reach}(t){\setminus} \{t\}: && 0  \leq p_s \leq 1  \\
    \forall s \in \textit{Reach}(t){\setminus} \{t\}: && \hspace*{-4ex}p_{s} \!=\! \!\!\sum_{a \in A(\!s\!)} \sum_{s' \in S}  p_{sa} {\cdot} \delta(s,a,s'){\cdot} p_{s'} \label{eq:ps}\\
    && p_{s_0}  \leq \gamma  \label{eq:gamma} \\
    \forall s \in S: && \Delta_s = \frac{1}{2}\!\!\!\sum_{a\in\Act(s)}|\sigma(s)(a){\shortminus} p_{sa}|\label{eq:delta_s}\\
    \forall s \in S: && 0\leq i_s\leq 1 \wedge \Delta_s \leq i_s \label{eq:counting}\\
    \forall s \in S: && D_0 = \sum_{s \in S} i_s \label{eq:d0}\\
    && D_1 = \frac{1}{|S|}\sum_{s\in S}\Delta_s\label{eq:d1}\\    
     \forall s \in S: &&  \Delta_{s} \leq D_\infty \label{eq:dinfty}
\end{eqnarray}

\noindent Here, Eq.\ (\ref{eq:min}) encodes the objective function value $d(\sigma,\sigma')$.
Constraints (\ref{eq:psa})-(\ref{eq:sum}) encode $p_{sa}$ as the probabilistic choices of $\sigma'$.
Constraints (\ref{eq:pt})-(\ref{eq:ps}) use the Bellman equations for computing the probabilities to reach $t$ from individual states, where $\textit{Reach}(t)$ is the set of all states from which $t$ is reachable (this can be easily computed by graph analysis).
Constraint (\ref{eq:gamma}) enforces that $Pr_{\mdp^{\sigma'}}(s_0, t) \leq \gamma$.
Finally, Constraints (\ref{eq:delta_s})-(\ref{eq:dinfty}) encode the distances $\textit{d}_{\bowtie}(\sigma, \sigma')$. Constraint \ref{eq:counting} ensures that a positive distance $\Delta_s$, indicating a difference between $\sigma(s)$ and $\sigma'(s)$, enforces $i_s=1$, and minimization will ensure $i_s=0$ for $\Delta_s=0$.
Note that for the infinity norm, even though Constraint \ref{eq:dinfty} only encodes that $D_\infty$ is an upper bound on the distribution distance $\Delta(\sigma(s),\sigma'(s))$ for all $s\in S$, minimizing the objective function will ensure that $D_\infty$ equals the smallest such value (i.e. the maximum) over all states.\looseness=-1


Note that the non-linearity of the problem stems from the calculation of $p_s$ in Constraint~(\ref{eq:ps}), since we allow probabilistic strategy decisions.

Let in the following $P$ denote the MIQCQP optimization problem defined by the Constraints (\ref{eq:min})-(\ref{eq:dinfty}). 

\begin{lemma}
\label{lemma:D}
Assume a solution to $P$, assigning to each variable $v$ the value $\nu(v)$. Let $\sigma'$ be the strategy for $\mdp$ with $\sigma'(s)(a)=\nu(p_{sa})$ for all $s\in S$ and $a\in\Act(s)$. Then the objective function value as specified in Constraint (\ref{eq:min}) equals $d(\sigma,\sigma')$.
\end{lemma}

\begin{proofsketch} 
We observe:
\begin{itemize}
\item $\Delta_s=\Delta(\sigma(s),\sigma'(s))$ according to Constraint (\ref{eq:delta_s});
\item $D_0=d_0(\sigma, \sigma')$ denotes the number of non-zero elements in $\Delta(\sigma, \sigma')$, using the counting mechanism from Eq. (\ref{eq:counting}). The variables $i_s$ indicate whether the entry $\Delta_s$ for $s$ in $\Delta(\sigma, \sigma')$ is non-zero. As $\Delta_s \leq 1$ holds for all $s \in S$, it follows that for $i_s =1$ we have $\Delta_s \leq i_s$.
    By minimizing $D_0$, each $i_s$ is set to $1$ if and only if $\Delta_s \neq 0$. 
\item $D_1=d_1(\sigma, \sigma')$ denotes the average over $\Delta(\sigma, \sigma')$.
\item    
    $D_\infty\,{=}\,d_\infty(\sigma, \sigma')$ encodes the maximal entry in $\Delta(\sigma, \sigma')$: by limiting each element $\Delta_s=\allowbreak\Delta(\sigma(s),\sigma'(s))$ from above by $D_\infty$ and by minimizing $D_\infty$, $D_\infty$ is forced to be the maximum.
\end{itemize}
Thus, the value of the objective function (\ref{eq:min}) is per definition $d(\sigma,\sigma')$.
\end{proofsketch}

\begin{theorem}[Soundness and Completeness]
$P$ admits a solution iff there exists a counterfactual strategy to $\sigma$ (under $d$ for reaching $t$ within $\gamma$ in $\mdp$).
\end{theorem}
\begin{proofsketch}
    $\rightarrow$. Let $\nu$ be a solution to $P$ assigning to each variable $v$ the value $\nu(v)$. 
    The values of the variables $p_{sa}$ induce a valid strategy $\sigma' \in \Sigma$ for $\mdp$, with $\sigma'(s)(a)=\nu(p_{sa})$ for all $s\in S$ and $a\in\Act(s)$.
    By satisfying (\ref{eq:gamma}), $\sigma'$ satisfies $Pr_{\mdp^{\sigma'}}(s_0, t) \leq \gamma$.
    According to Lemma \ref{lemma:D}, the objective function value is $d(\sigma,\sigma')$. As the solution minimizes the objective function value, there exists no strategy with smaller distance.
    Hence, $\sigma'$ is a counterfactual strategy to $\sigma$ (under $d$ for reaching $t$ within $\gamma$ in $\mdp$).

    $\leftarrow$.
Let $\sigma' \in \Sigma$ be a counterfactual strategy to $\sigma$. The strategy can be extended to a solution for the optimization problem by (1) encoding $\sigma'$ into variables $p_{sa} = \sigma'(s)(a)$, (2) computing reachability values for $p_s$ satisfying the Bellman optimality equation, (3) setting $i_s = 1$ iff any decision in state $s$ was changed, and (4) setting $\Delta_s$ and the distance variables $D_0$, $D_1$ and $D_{\infty}$ accordingly.
    As $\sigma'$ is well-defined, all constraints in $P$ are satisfied, and from Definition~\ref{def:cs} it follows that $\sigma'$ minimizes $d(\sigma,\sigma')$.
\end{proofsketch}

The following theorem prohibits efficient, i.e. polynomial-time, algorithms for solving the MIQCQP optimization problem for counterfactual strategies.
\begin{theorem}
    The presented optimization problem for counterfactual strategies is generally nonconvex.
\end{theorem}
\begin{proof}
    Recall that an optimization problem is convex iff the target function and all constraints are convex~\cite{boyd2004convex}.
    We show that the constraints of our optimization problem are, in general, not convex by providing a counterexample:
    Consider the MDP defined in Fig.~\ref{fig:loan-example}.
    The quadratic constraint stemming from~(\ref{eq:ps}) for encoding $p_{s_0}$ can be expressed as follows:
    \begin{multline*}
    p_{s_0} = p_{{s_0}{\text{Apply}}}  \cdot 0.95\cdot p_{\text{Application}} + p_{{s_0}{\text{Apply}}} \cdot 0.05 \cdot p_{\text{Error}} + \\ p_{{s_0}{\text{Consult}}}  \cdot 1 \cdot p_{\text{Consultation}} = \\ 
    \begin{pmatrix}
    p_{{s_0}{\text{Apply}}} \\
    p_{\text{Application}} \\
    p_{\text{Error}} \\
    p_{{s_0}{\text{Consult}}} \\
    p_{\text{Consultation}} \\
    \end{pmatrix}^T
    \underbrace{\begin{pmatrix}
        0 & \frac{0.95}{2} & 0 & 0 & 0 \\
        \frac{0.95}{2} & 0 & \frac{0.05}{2} & 0 & 0 \\
        0 & \frac{0.05}{2} & 0 & 0 & 0 \\
        0 & 0 & 0 & 0 & \frac{1}{2} \\
        0 & 0 & 0 &\frac{1}{2} & 0 \\
    \end{pmatrix}}_{P_{s_0}}
    \begin{pmatrix}
    p_{{s_0}{\text{Apply}}} \\
    p_{\text{Application}} \\
    p_{\text{Error}} \\
    p_{{s_0}{\text{Consult}}} \\
    p_{\text{Consultation}} \\
    \end{pmatrix}
    \end{multline*}
    Our goal is now to check whether the function $f \colon \R^5 \to \R$ defined by $f(x) = x^T P_{s_0} x$ for $x \in \R^5$ is convex.
    Oberserve that the Hessian of $x^T P_{s_0} x$ is $2P_{s_0}$.
    The eigenvalues of $2P_{s_0}$ are $-1$, $1$, $-\frac{\sqrt{362}}{20}$, $\frac{\sqrt{362}}{20}$, and $0$.
    As the matrix is symmetric and has a negative eigenvalue, it is not positive semi-definite.
    By the second-order condition of convexity~\cite{boyd2004convex}, the constraint is thus not convex, and the whole problem is neither.
\end{proof}

\paragraph{Remark.} We note that validity, actionability, proximity, and sparsity are ensured by construction in our approach. \emph{Validity} of counterfactual strategies requires that $\Pr_{\mdp^\sigma}(s_0, t) \leq \gamma$, i.e. following the counterfactual strategy $\sigma'$ reduces the chance of reaching $t$ below the limit $\gamma$, ensured by Constraint (\ref{eq:gamma}).
%
\emph{Proximity} minimizes the changes in the counterfactual strategy $\sigma'$.
The minimization of the strategy distance measure $d$
ensures that $\sigma'$ is as close to $\sigma$ as possible but yet satisfies $\Pr_{\mdp^\sigma}(s_0, t) \leq \gamma$.
\emph{Actionability} of counterfactual strategies follows from a valid MDP where only actually controllable features are controllable.
%
\emph{Sparsity} between initial strategy $\sigma$ and counterfactual strategy $\sigma'$ follows from minimizing the strategy distance measure.


\paragraph{Diverse Counterfactual Strategies.}
A single counterfactual strategy provides only a single alternative, e.g., for recourse.
However, recent work~\cite{bove2023investigating} has shown that offering diverse counterfactuals demonstrating different possibilities for recourse may improve the interpretability of AI decisions. 
%
To this end, we extend our method for computing an individual counterfactual strategy to an iterative method where the $n$th counterfactual minimizes the distance to the initial strategy while maximizing the distance to all previously generated solutions.\looseness=-1

We define the diversity of a collection of strategies $\sigma_0, \dots, \sigma_n \in \Sigma$ as the determinant of the matrix of inverse pairwise distances $D \in \R^{n \times n}$ with $D_{ij} = \frac{1}{1+\|\Delta(\sigma_i, \sigma_j')\|_1}$, as done in~\cite{mothilal2020explaining}.

Further, we adjust the objective function (\ref{eq:min}) to additionally optimize for diversity:
\[
    \min
    r_0 \cdot D_0 + r_1 \cdot D_1 + r_\infty \cdot D_\infty  - \lambda \cdot det(D) .
\]

To avoid ill-defined determinants, a small perturbation is added to
the diagonals.  In comparison to~\cite{mothilal2020explaining}, we do
not average the $\| \cdot \|_1$ norm as each element is smaller than
$1$ and we wish to maximize for diversity.  The parameter $\lambda$
weights the diversity part of the target function.  In this work,
we use $r_0 = r_1 = r_{\infty} = 1$ and $\lambda = 2$ to weight
each distance component equally and to weight diversity
higher than distances.

To evaluate the diversity of counterfactual strategies, we consider
the fraction of novel state-action pairs introduced.  By this, a
diverse counterfactual strategy contains many state-action pairs not
utilized in previously.



\section{Experimental Evaluation}
\label{sec:experiments}
Our aim is to
demonstrate 
that (1) counterfactual strategies can be efficiently computed for
complex sequential decision processes and
(2)
diverse counterfactual strategies can be generated.
This is done by experiments \textbf{Exp1} and \textbf{Exp2},
respectively, conducted on complementary real-world datasets.\looseness=-1

\subsection{Experimental Design and Setup}

In our experiments, we consider
four  real-world datasets. GrepS records customer interaction with
a
programming skill evaluation service~\cite{kobialkaWeightedGames}.
BPIC12~\cite{bpi2012} and BPIC17~\cite{bpi2017}, which record the loan
application procedure in a
bank, stem from 
the
\emph{Business Process Intelligence
  Challenge}\footnote{\url{https://www.tf-pm.org/competitions-awards/bpi-challenge}}
of the IEEE Task Force on Process
Mining.\footnote{\url{https://www.tf-pm.org/}}
%
MSSD is the \emph{Music Streaming Sessions
  Dataset}~\cite{brost2019music} from
Spotify;
we consider the small version of MSSD, with
$10\,000$ listening sessions.

We briefly outline the experimental setup (for further details, see
Appendix~\ref{appendix}).  After standard preprocessing of the
datasets,
stochastic automata learning~\cite{DBLP:journals/ml/MaoCJNLN16} was
used to generate the MDPs.  Given the size of the MSSD dataset, we
construct 10 models depending on the number of included traces; e.g.,
MSSD10 and MSSD40 include 10\% and 40\% of the data set, respectively.
For MSSD, the number of states in each model is drastically higher
than for the other datasets: already MSSD10 has on average $40$
times more states than BPIC12 or BPIC17, and a $100$ times higher
$\max$ degree.  We randomly generate ten initial user strategies for
each model and let the target probability $\gamma$ range over
$\{0.0001\} \cup \{0.1, 0.2, \dots, 1 \}$, where $0.0001$ represents
near-perfect performance.

\subsection{Computing Counterfactual Strategies}
In \textbf{Exp1}, we compute
counterfactual strategies for all models, showing that counterfactual
strategies for models with thousands of states and tens of thousands
of transitions can be computed within minutes.

Table~\ref{tab:greps_bpic} compares averaged computation times and
outcomes for all values of $\gamma$ for GrepS,
BPIC12 and BPIC17; Opt.\ denotes optimally solved
instances, Inf.\ infeasible instances and T.O.\
timeouts.
No computation took more
than a minute, e.g.\ for
BPIC17 the longest computation took $5.01$ seconds.
The mean for all models is around one second.

\begin{figure}[t]
  \centering
\includegraphics[trim={6pt 8 6 6},clip,width=0.8\linewidth]{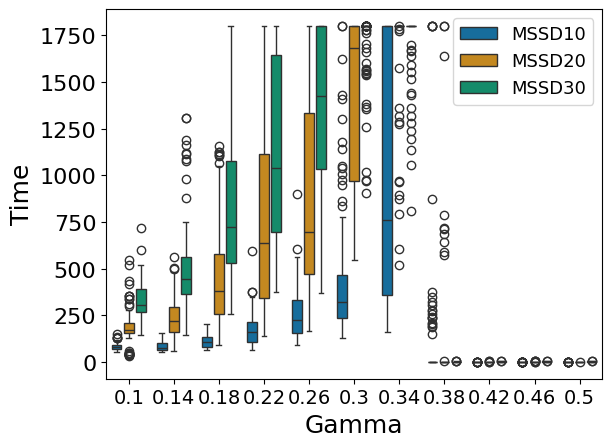}
    \caption{Runtime comparison.}
    \label{fig:mssd_detailed}
\end{figure}

Table~\ref{tab:mssd} shows
runtimes for the MSSD models;
Sub.O.\ denotes
instances
solved within the time limit, but not necessarily optimally.  While
MSSD10--MSSD30 had few timeouts, MSSD70--MSSD100 had $269$--$300$
timeouts.  Table~\ref{tab:mssd_quantiles_short} shows individual
results for sextile 1 (S1) to sextile 3 (S3) of $\gamma$; the full
table is
in Appendix~\ref{appendix}.  For all MSSD models both trivial and
infeasible problems are solved, see S1 and S3 in
Table~\ref{tab:mssd_sextiles_category}.
Figure~\ref{fig:mssd_detailed} details runtimes for MSSD10--MSSD30
with $\gamma \in [0.1, 0.5]$, highlighting the runtime peak for
non-trivial instances.  The difficulty lies in computing non-trivial
counterfactual strategies around S2, where all non-trivial models from
MSSD50--MSSD100 timeout, see Table~\ref{tab:mssd_sextiles_runtime}.
\looseness=-1


\begin{table}
    \centering
    \small
    \caption{Averaged GrepS and BPIC runtime results in seconds for computing counterfactual strategies.}
    \label{tab:greps_bpic}
    \setlength{\tabcolsep}{2pt}
    \begin{tabular}{lrrrrrrr}
    \toprule
    \textbf{Model} &  \textbf{mean(t)} &  \textbf{std(t)} &  \textbf{min(t)} &  \textbf{max(t)} &  \textbf{Opt.} &  \textbf{Inf.} &  \textbf{T.O.} \\
    \midrule
    \textbf{Greps  } &     0.01 &    0.01 &    0.01 &    0.05 &    90 &    20 &     0 \\
    \textbf{BPIC12} &     0.79 &    0.94 &    0.04 &    4.24 &   110 &     0 &     0 \\
    \textbf{BPIC17} &     1.00 &    1.06 &    0.01 &    5.01 &   100 &    10 &     0 \\
    \bottomrule

    \end{tabular}
\end{table} 

\begin{table}
    \centering
    \caption{MSSD runtime results in seconds for computing counterfactual strategies.}
    \small
    \setlength{\tabcolsep}{1pt}
    \label{tab:mssd}
    \begin{tabular}{lrrrrrrrrr}
    \toprule
        \textbf{Model} &  \textbf{mean(t)} &  \textbf{std(t)} &  \textbf{min(t)} &  \textbf{max(t)} &  \textbf{Opt.} &  \textbf{Inf.} &  \textbf{T.O.} &  \textbf{Sub.O.} \\
        \midrule
        \textbf{MSSD10}  &    56.08 &  152.35 &    0.05 &   T.O. &   710 &   388 &     2 &       0 \\
        \textbf{MSSD20}  &   195.77 &  453.00 &    0.10 &   T.O. &   707 &   343 &    41 &       9 \\
        \textbf{MSSD30}  &   276.62 &  556.85 &    0.14 &   T.O. &   703 &   320 &    60 &      17 \\
        \textbf{MSSD40}  &   412.96 &  718.68 &    0.18 &   T.O. &   700 &   193 &   207 &       0 \\
        \textbf{MSSD50}  &   464.38 &  767.34 &    0.22 &   T.O. &   700 &   149 &   251 &       0 \\
        \textbf{MSSD60}  &   483.53 &  788.52 &    0.24 &   T.O. &   700 &   123 &   277 &       0 \\
        \textbf{MSSD70}  &   485.84 &  789.70 &    0.29 &   T.O. &   700 &   131 &   269 &       0 \\
        \textbf{MSSD80}  &   493.91 &  799.60 &    0.33 &   T.O. &   700 &   103 &   297 &       0 \\
        \textbf{MSSD90}  &   494.57 &  799.79 &    0.36 &   T.O. &   700 &   100 &   300 &       0 \\
        \textbf{MSSD100} &   494.39 &  799.90 &    0.39 &   T.O. &   700 &   100 &   300 &       0 \\
        \bottomrule
    \end{tabular}
\end{table}

\begin{table*}
    \scriptsize
    \caption{MSSD results for selected sextiles of $\gamma$.}
    \label{tab:mssd_quantiles_short}
    \setlength{\tabcolsep}{1pt}
    \begin{subtable}[t]{0.42\linewidth}
        \centering
        \begin{tabular}{lrrrrrrrrrrrr}
            \toprule
            & \multicolumn{4}{c}{\textbf{S1 (0, 0.17]}} & \multicolumn{4}{c}{\textbf{S2 (0.17, 0.33]}} & \multicolumn{4}{c}{\textbf{S3 (0.33, 0.5]}} \\
            {Model} &  Opt. &    Inf. &    T.O. & Sub. &    Opt. &    Inf. &    T.O. &  Sub. &  Opt. &    Inf. &    T.O. &  Sub. \\
            \midrule
            \textbf{MSSD10 } &       0 &        200 &       0 &          0 &      10 &        188 &       2 &          0 &     200 &          0 &       0 &          0 \\
            \textbf{MSSD20 } &       0 &        200 &       0 &          0 &       7 &        143 &      41 &          9 &     200 &          0 &       0 &          0 \\
            \textbf{MSSD30 } &       0 &        200 &       0 &          0 &       3 &        120 &      60 &         17 &     200 &          0 &       0 &          0 \\
            \textbf{MSSD40 } &       0 &        187 &      13 &          0 &       0 &          6 &     194 &          0 &     200 &          0 &       0 &          0 \\
            \textbf{MSSD50 } &       0 &        149 &      51 &          0 &       0 &          0 &     200 &          0 &     200 &          0 &       0 &          0 \\
            \textbf{MSSD60 } &       0 &        123 &      77 &          0 &       0 &          0 &     200 &          0 &     200 &          0 &       0 &          0 \\
            \textbf{MSSD70 } &       0 &        131 &      69 &          0 &       0 &          0 &     200 &          0 &     200 &          0 &       0 &          0 \\
            \textbf{MSSD80 } &       0 &        103 &      97 &          0 &       0 &          0 &     200 &          0 &     200 &          0 &       0 &          0 \\
            \textbf{MSSD90 } &       0 &        100 &     100 &          0 &       0 &          0 &     200 &          0 &     200 &          0 &       0 &          0 \\
            \textbf{MSSD100} &       0 &        100 &     100 &          0 &       0 &          0 &     200 &          0 &     200 &          0 &       0 &          0 \\
            \bottomrule
        \end{tabular}
        \caption{Categorical results by sextile.}
        \label{tab:mssd_sextiles_category}
    \end{subtable}\qquad
    \begin{subtable}[t]{0.48\linewidth}
        \centering
        \begin{tabular}{lrrrrrrrrrrrr}
            \toprule
            & \multicolumn{4}{c}{\textbf{S1 (0, 0.17]}} & \multicolumn{4}{c}{\textbf{S2 (0.17, 0.33]}} & \multicolumn{4}{c}{\textbf{S3 (0.33, 0.5]}} \\
            {Model} & mean(t) &  std(t) & min(t) &  max(t) & mean(t) &  std(t) & min(t) &  max(t) & mean(t) & std(t) & min(t) & max(t) \\
            \midrule
            \textbf{MSSD10 } &   39 &  40 &   0 &    128 &     266 & 267 &       73 &    T.O. &      1 &    0 &   1 &   2 \\
            \textbf{MSSD20 } &   92 & 114 &   0 &    684 &     978 & 603 &      126 &    T.O. &      2 &    0 &   1 &   4 \\
            \textbf{MSSD30 } &  159 & 167 &   0 &    567 &    1352 & 495 &      335 &    T.O. &      3 &    1 &   2 &   5 \\
            \textbf{MSSD40 } &  468 & 585 &   0 &   T.O. &    1793 &  59 &     1144 &    T.O. &      3 &    1 &   2 &   7 \\
            \textbf{MSSD50 } &  743 & 796 &   0 &   T.O. &    T.O. &   0 &     T.O. &    T.O. &      3 &    1 &   2 &   6 \\
            \textbf{MSSD60 } &  847 & 865 &   0 &   T.O. &    T.O. &   0 &     T.O. &    T.O. &      3 &    1 &   2 &   8 \\
            \textbf{MSSD70 } &  858 & 867 &   0 &   T.O. &    T.O. &   0 &     T.O. &    T.O. &      4 &    1 &   3 &   7 \\
            \textbf{MSSD80 } &  899 & 901 &   0 &   T.O. &    T.O. &   0 &     T.O. &    T.O. &      5 &    1 &   3 &  13 \\
            \textbf{MSSD90 } &  900 & 902 &   0 &   T.O. &    T.O. &   0 &     T.O. &    T.O. &      6 &    1 &   3 &  12 \\
            \textbf{MSSD100} &  900 & 902 &   0 &   T.O. &    T.O. &   0 &     T.O. &    T.O. &      5 &    1 &   4 &  12 \\            
            \bottomrule
        \end{tabular}
        \caption{Runtime results by sextile, rounded to seconds.}
        \label{tab:mssd_sextiles_runtime}
    \end{subtable}
\end{table*}


The counterfactual strategies can be used to provide interpretable
recommendations to users, which is essential to enable users to follow
the recourse suggested by the counterfactual strategy.
To this aim,
counterfactual strategies are presented in a textual
representation highlighting the suggested changes.  An example for two states of BPIC12 is given below,
where the client is asked
not to cancel the loan offer after receiving the first offer but to
continue in the process:

\begin{scriptsize}
\begin{verbatim}
State `negative' is reached with probability 0.64.
You can reach `negative' with probability 0.09 as follows:
 In state `q9: A_CANCELLED' 
  increase probability of action `Nabellen offer.' to 0.89
  decrease probability of action `negative' to 0.07
 In state `q27: Nabellen offer.#0' 
  increase probability of action `O_SENT_BACK' to 0.81
  decrease probability of action `negative' to 0.0
  decrease probability of action `O_CANCELLED' to 0.0
\end{verbatim}
\end{scriptsize}

We summarize our conclusions
for \textbf{Exp1}: counterfactual strategies can be efficiently
computed for complex MDPs with up to $10\,000$ states and
$20\,000$ transitions; these models are significantly larger than
models used in current process mining benchmarks but occur in, e.g.,
MSSD.

\begin{figure}[t]
\centering

\begin{subfigure}[t]{0.8\linewidth}
  \centering
        \includegraphics[trim={10pt 10 5 5},clip,width=0.9\linewidth]{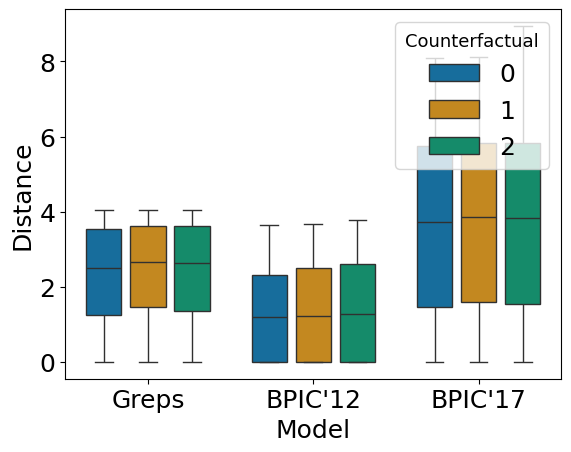}
        \caption{Boxplot over distances from three diverse counterfactual strategies to the initial strategy.}
        \label{fig:diversity_agg_value}
    \end{subfigure}
    \begin{subfigure}[t]{0.8\linewidth}
    \centering
        \includegraphics[trim={1pt 10 5 5},clip,width=0.9\linewidth]{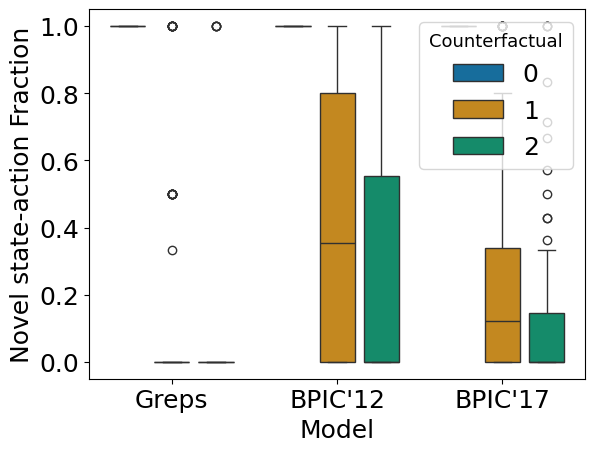}
        \caption{Boxplot showing novel state-action fractions for three diverse counterfactual strategies.}
        \label{fig:diverse_agg_unknown_fraction}
    \end{subfigure}
    \caption{Results for diverse counterfactual strategies.}
    \label{fig:diverse_results}
\end{figure}

\subsection{Diverse Counterfactuals}
In \textbf{Exp2}, we generate a collection of diverse counterfactual
strategies for GrepS, BPIC12, and BPIC17 and compare the distance
between counterfactual strategies as well as the diversity
of the counterfactual strategies.
To evaluate diversity,
we investigate the fraction of novel state-action pairs, i.e.,
the actions changed in a counterfactual strategy compared to the
initial strategy that were not changed in any previous one.

Figure~\ref{fig:diverse_results} compares the distance between each
generated counterfactual strategy and the initial
strategy (Fig.~\ref{fig:diversity_agg_value}), and shows the fraction
of novel state-action pairs
in each strategy (Fig.~\ref{fig:diverse_agg_unknown_fraction}).  While
the distance to the initial strategy varies only slightly between
consecutive counterfactual strategies, each provides novel recourse
strategies.  Intermediate values of $\gamma$
offer the largest range of diversity for counterfactual
explanations.  The individual distances to the initial strategy and the
fraction of novel state-action pairs increase with $\gamma$, until the
problem is trivially satisfied, see Fig.~\ref{fig:div_ce_gamma} in
Appendix~\ref{appendix}.

We summarize our conclusions
for \textbf{Exp2}: diverse counterfactual strategies can be
efficiently computed for the benchmark problems considered. The
diverse counterfactual strategies introduce new recourse possibilities
while remaining at a short distance to the initial strategy,
comparable to the distance of the first counterfactual strategy.


\section{Related Work}
\label{sec:related_work}
We discuss related work with respect to stochastic counterfactuals,
model repair and synthesis.  Balke and Pearl discuss stochastic
evaluations of counterfactual queries in their seminal
work~\cite{balke1994probabilistic}.  Since then, much work on
counterfactual explanations has been published, summarized by
Guidotti~\cite{guidotti2024counterfactual}.  An adaptation for
predictive business process monitoring was presented with
\emph{LORLEY}~\cite{huang2021counterfactual}.  Notably, MDPs in
combination with causal models have been adapted for counterfactual
reasoning~\cite{tsirtsis2021counterfactual,kazemi2024counterfactual}.
In this line of work, the authors define the transition probabilities
of the MDP via structural causal models and then search for a
counterfactual path that only diverges in $k$ actions from the given
path.  In contrast, our work does not assume causal knowledge. We
further define counterfactuals over strategies, as opposed to
paths in the MDP.

Although both model repair and counterfactuals  start with a model and a violated property, these problems aim for different solutions.
Model repair considers the problem of adjusting a model to satisfy a desired property~\cite{bartocci2011model,chatzieleftheriou2018abstract,chen2013model,pathak2015greedy}. 
In contrast, counterfactual strategies do not aim to adjust the underlying model, but rather to propose behavior changes to the user (thus, the transition probabilities of the model remain unchanged).

Recent work on model synthesis for parametric MDPs consider
transition probabilities
expressed as functions over
variables.  In this setting, one searches for a parameter valuation
such that a given property is satisfied under every
strategy~\cite{cubuktepe2017sequential,cubuktepe2018synthesis,cubuktepe2021convex}.
In contrast, our work starts from a fully specified model and a
strategy, and we search for a minimal change of the strategy that
satisfies the given property.\looseness=-1


\section{Conclusion}
\label{sec:conclusion}
In this work, we introduced counterfactual strategies for Markov decision processes as post-hoc explanations for sequential decision-making tasks. 
We presented an optimization approach for computing counterfactual strategies and extended it to also optimize for diversity. In extensive experiments on four real-world datasets, we evaluated the generation of diverse counterfactual strategies, showing that counterfactual strategies can be generated within minutes for models significantly larger than current Process Mining benchmarks.

Our work opens several interesting avenues for future work. First, we plan to investigate the complexity of generating counterfactual strategies further, as well as techniques to further reduce the runtime of our approach. Approximations of the optimization problem, 
such as those based on linearization, e.g.,~\cite{cubuktepe2021convex}, 
promise to reduce the computation time while producing local optimal strategies. Second, it would be interesting to study the problem of generating  counterfactual strategies for scenarios where the environment (e.g., the service provider) adapts to counterfactual changes. This will require generalizing our counterfactual strategies from MDPs to Stochastic Games, thus requiring novel theoretical investigations. Finally, it would be interesting to investigate whether our counterfactual strategies result in realistic recourse behaviors, e.g. by measuring similarity to those witnessed in our dataset or by running user studies.


\section*{Acknowledgments}
Leofante was supported by Imperial College through the Imperial
College Research Fellowship scheme.  Kobialka, Tapia Tarifa, and
Johnsen were supported by the \emph{Smart Journey Mining} project,
funded by the Research Council of Norway (grant no.\,312198).  Gerlach
is supported by the DFG RTG 2236/2 project \emph{UnRAVeL}.

\bibliographystyle{named}
\bibliography{references}

\clearpage
\appendix
\section{Appendix}\label{appendix}
\label{sec:appendix}

\begin{table}
    \centering
    \caption{Averaged model overview of each dataset.}
    \label{tab:model_overview}
    \setlength{\tabcolsep}{2pt}
    \begin{tabular}{lrrr}
        \toprule
        {} &  $mean(\text{States})$ &  $mean(\text{Transitions})$ &  $max$ Degree \\
        \midrule
        \textbf{Greps      } &    64.0 &         76.0 &         6.0 \\
        \textbf{BPIC'12    } &   129.0 &        288.0 &        28.0 \\
        \textbf{BPIC'17} &   158.0 &        292.0 &        19.0 \\
        \textbf{MSSD10     } &  5702.8 &      11341.5 &      1952.6 \\
        \textbf{MSSD20     } &  7971.6 &      16247.5 &      2794.0 \\
        \textbf{MSSD30     } &  9378.6 &      19718.0 &      3397.3 \\
        \textbf{MSSD40     } & 10603.4 &      22585.8 &      3857.1 \\
        \textbf{MSSD50     } & 11581.2 &      25016.5 &      4249.0 \\
        \textbf{MSSD60     } & 12456.2 &      27106.1 &      4612.0 \\
        \textbf{MSSD70     } & 13213.2 &      28984.2 &      4918.7 \\
        \textbf{MSSD80     } & 13804.2 &      30547.4 &      5171.3 \\
        \textbf{MSSD90     } & 14334.0 &      31905.5 &      5403.8 \\
        \textbf{MSSD100    } & 14860.0 &      33266.0 &      5622.0 \\
        \bottomrule
        \end{tabular}
\end{table}

\begin{table*}
    \centering
    \caption{GrepS and BPIC runtime results in seconds for quantiles.}
    \label{tab:greps_bpic_quantiles}
    \setlength{\tabcolsep}{2pt}
    \begin{tabular}{lrrrrrrrrrrrrrrrr}
    \toprule
    & \multicolumn{4}{c}{\textbf{Q1 (0, 0.25]}} & \multicolumn{4}{c}{\textbf{Q2 (0.25, 0.5]}} & \multicolumn{4}{c}{\textbf{Q3 (0.5, 0.75]}} &  \multicolumn{4}{c}{\textbf{Q4 (0.75, 1]}} \\
    Model & mean(t) & std(t) & min(t) & max(t) & mean(t) & std(t) & min(t) & max(t) & mean(t) & std(t) & min(t) & max(t) & mean(t) & std(t) & min(t) & max(t) \\
    \midrule
    \textbf{Greps  } &     0.0 &    0.0 &    0.0 &    0.0 &     0.0 &    0.0 &    0.0 &    0.0 &     0.0 &    0.0 &    0.0 &    0.0 &     0.0 &    0.0 &    0.0 &    0.1 \\
    \textbf{BPIC'12} &     1.3 &    0.8 &    0.4 &    3.3 &     1.2 &    1.1 &    0.0 &    4.2 &     0.3 &    0.5 &    0.0 &    1.5 &     0.1 &    0.2 &    0.0 &    1.0 \\
    \textbf{BPIC'17} &     1.4 &    1.5 &    0.0 &    5.0 &     1.5 &    0.8 &    0.6 &    3.8 &     0.8 &    0.5 &    0.4 &    2.2 &     0.2 &    0.2 &    0.0 &    0.8 \\
    \bottomrule
    \end{tabular}
\end{table*}


\begin{table*}
    \centering
    \scriptsize
    \caption{MSSD results by sextiles.}
    \label{tab:mssd_quantiles}
    \setlength{\tabcolsep}{1pt}
    \begin{subtable}[t]{\linewidth}
        \centering
        \begin{tabular}{lrrrrrrrrrrrrrrrrrrrrrrrr}
            \toprule
            & \multicolumn{4}{c}{\textbf{S1 (0, 0.17]}} & \multicolumn{4}{c}{\textbf{S2 (0.17, 0.33]}} & \multicolumn{4}{c}{\textbf{S3 (0.33, 0.5]}} &  \multicolumn{4}{c}{\textbf{S4 (0.5, 0.67]}} & \multicolumn{4}{c}{\textbf{S5 (0.67, 0.83]}} & \multicolumn{4}{c}{\textbf{S6 (0.83, 1]}}  \\
            {Model} &  Opt. &    Inf. &    T.O. & Sub. &    Opt. &    Inf. &    T.O. &  Sub. &  Opt. &    Inf. &    T.O. &  Sub. &  Opt. &    Inf. &    T.O.  & Sub. & Opt. &    Inf. &    T.O. & Sub. & Opt. &    Inf. &    T.O. & Sub.  \\
            \midrule
            \textbf{MSSD10 } &       0 &        200 &       0 &          0 &      10 &        188 &       2 &          0 &     200 &          0 &       0 &          0 &     100 &          0 &       0 &          0 &     200 &          0 &       0 &          0 &     200 &          0 &       0 &          0 \\
            \textbf{MSSD20 } &       0 &        200 &       0 &          0 &       7 &        143 &      41 &          9 &     200 &          0 &       0 &          0 &     100 &          0 &       0 &          0 &     200 &          0 &       0 &          0 &     200 &          0 &       0 &          0 \\
            \textbf{MSSD30 } &       0 &        200 &       0 &          0 &       3 &        120 &      60 &         17 &     200 &          0 &       0 &          0 &     100 &          0 &       0 &          0 &     200 &          0 &       0 &          0 &     200 &          0 &       0 &          0 \\
            \textbf{MSSD40 } &       0 &        187 &      13 &          0 &       0 &          6 &     194 &          0 &     200 &          0 &       0 &          0 &     100 &          0 &       0 &          0 &     200 &          0 &       0 &          0 &     200 &          0 &       0 &          0 \\
            \textbf{MSSD50 } &       0 &        149 &      51 &          0 &       0 &          0 &     200 &          0 &     200 &          0 &       0 &          0 &     100 &          0 &       0 &          0 &     200 &          0 &       0 &          0 &     200 &          0 &       0 &          0 \\
            \textbf{MSSD60 } &       0 &        123 &      77 &          0 &       0 &          0 &     200 &          0 &     200 &          0 &       0 &          0 &     100 &          0 &       0 &          0 &     200 &          0 &       0 &          0 &     200 &          0 &       0 &          0 \\
            \textbf{MSSD70 } &       0 &        131 &      69 &          0 &       0 &          0 &     200 &          0 &     200 &          0 &       0 &          0 &     100 &          0 &       0 &          0 &     200 &          0 &       0 &          0 &     200 &          0 &       0 &          0 \\
            \textbf{MSSD80 } &       0 &        103 &      97 &          0 &       0 &          0 &     200 &          0 &     200 &          0 &       0 &          0 &     100 &          0 &       0 &          0 &     200 &          0 &       0 &          0 &     200 &          0 &       0 &          0 \\
            \textbf{MSSD90 } &       0 &        100 &     100 &          0 &       0 &          0 &     200 &          0 &     200 &          0 &       0 &          0 &     100 &          0 &       0 &          0 &     200 &          0 &       0 &          0 &     200 &          0 &       0 &          0 \\
            \textbf{MSSD100} &       0 &        100 &     100 &          0 &       0 &          0 &     200 &          0 &     200 &          0 &       0 &          0 &     100 &          0 &       0 &          0 &     200 &          0 &       0 &          0 &     200 &          0 &       0 &          0 \\
            \bottomrule
        \end{tabular}
        \caption{Categorical results by sextile.}
    \end{subtable}

    \begin{subtable}[t]{\linewidth}
        \centering
        \begin{tabular}{lrrrrrrrrrrrrrrrrrrrrrrrr}
            \toprule
            & \multicolumn{4}{c}{\textbf{S1 (0, 0.17]}} & \multicolumn{4}{c}{\textbf{S2 (0.17, 0.33]}} & \multicolumn{4}{c}{\textbf{S3 (0.33, 0.5]}} &  \multicolumn{4}{c}{\textbf{S4 (0.5, 0.67]}} & \multicolumn{4}{c}{\textbf{S5 (0.67, 0.83]}} & \multicolumn{4}{c}{\textbf{S6 (0.83, 1]}}  \\
            {Model} & mean(t) &  std(t) & min(t) &  max(t) & mean(t) &  std(t) & min(t) &  max(t) & mean(t) & std(t) & min(t) & max(t) & mean(t) & std(t) & min(t) &  max(t) & mean(t) &  std(t) & min(t) &  max(t) & mean(t) &  std(t) & min(t) &  max(t) \\
            \midrule
            \textbf{MSSD10 } &   39 &  40 &   0 & 128 &  266 & 267 &   73 &   T.O. &    1 &   0 &   1 &   2 &    1 &   0 &   1 &   2 &    1 &   0 &   1 &   2 &    1 &   0 &   1 &   2 \\
            \textbf{MSSD20 } &   92 & 114 &   0 & 684 &  978 & 603 &  126 &   T.O. &    2 &   0 &   1 &   4 &    2 &   0 &   2 &   3 &    2 &   0 &   2 &   4 &    2 &   0 &   2 &   4 \\
            \textbf{MSSD30 } &  159 & 167 &   0 & 567 & 1352 & 495 &  335 &   T.O. &    3 &   1 &   2 &   5 &    3 &   1 &   2 &   6 &    3 &   0 &   2 &   6 &    3 &   0 &   2 &   5 \\
            \textbf{MSSD40 } &  468 & 585 &   0 &   T.O. & 1793 &  59 & 1144 &   T.O. &    3 &   1 &   2 &   7 &    3 &   1 &   2 &   7 &    3 &   1 &   2 &   7 &    3 &   1 &   2 &   7 \\
            \textbf{MSSD50 } &  743 & 796 &   0 &   T.O. &    T.O. &   0 &    T.O. &   T.O. &    3 &   1 &   2 &   6 &    3 &   1 &   2 &   5 &    3 &   1 &   2 &   8 &    3 &   1 &   2 &   6 \\
            \textbf{MSSD60 } &  847 & 865 &   0 &   T.O. &    T.O. &   0 &    T.O. &   T.O. &    3 &   1 &   2 &   8 &    3 &   1 &   2 &   6 &    3 &   1 &   2 &  10 &    3 &   1 &   2 &   9 \\
            \textbf{MSSD70 } &  858 & 867 &   0 &   T.O. &    T.O. &   0 &    T.O. &   T.O. &    4 &   1 &   3 &   7 &    4 &   1 &   3 &   7 &    4 &   1 &   3 &   7 &    4 &   1 &   3 &   9 \\
            \textbf{MSSD80 } &  899 & 901 &   0 &   T.O. &    T.O. &   0 &    T.O. &   T.O. &    5 &   1 &   3 &  13 &    5 &   1 &   4 &  14 &    5 &   1 &   3 &  11 &    5 &   1 &   3 &   8 \\
            \textbf{MSSD90 } &  900 & 902 &   0 &   T.O. &    T.O. &   0 &    T.O. &   T.O. &    6 &   1 &   3 &  12 &    6 &   2 &   4 &  14 &    5 &   1 &   3 &  11 &    6 &   1 &   4 &  11 \\
            \textbf{MSSD100} &  900 & 902 &   0 &   T.O. &    T.O. &   0 &    T.O. &   T.O. &    5 &   1 &   4 &  12 &    6 &   1 &   4 &  10 &    5 &   1 &   4 &  11 &    5 &   1 &   4 &  12 \\
            \bottomrule
        \end{tabular}
        \caption{Runtime results by sextile, rounded to seconds.}
    \end{subtable}
\end{table*}

\begin{figure*}
    \begin{subfigure}[t]{0.3\linewidth}
        \includegraphics[width=\linewidth]{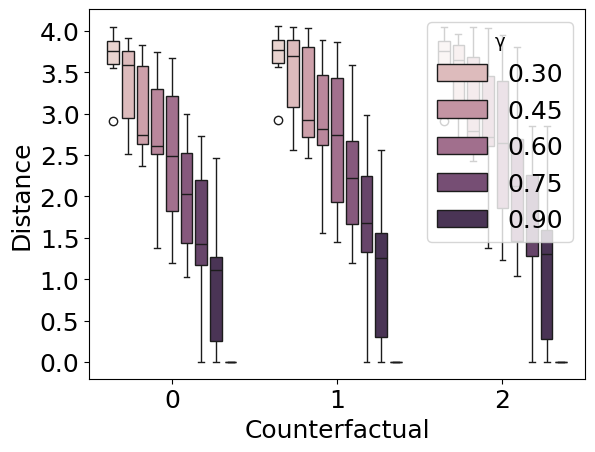}
        \caption{Greps.}
    \end{subfigure}
    \begin{subfigure}[t]{0.3\linewidth}
        \includegraphics[width=\linewidth]{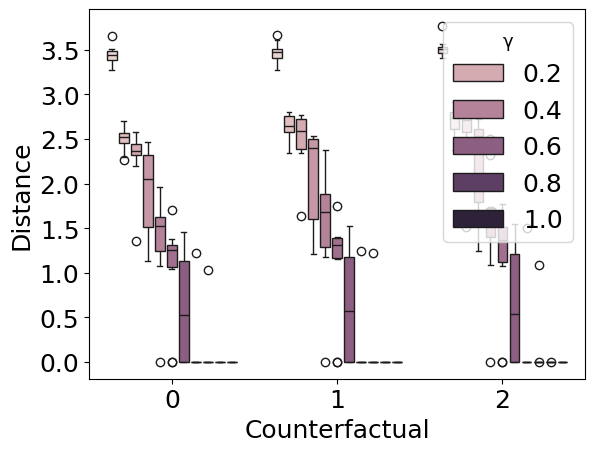}
        \caption{BPIC'12.}
    \end{subfigure}
    \begin{subfigure}[t]{0.3\linewidth}
        \includegraphics[width=\linewidth]{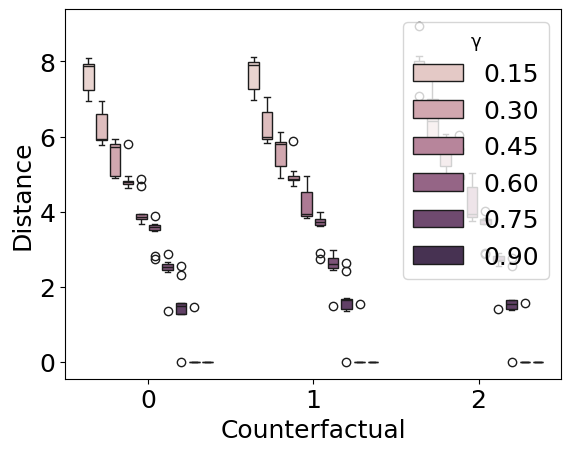}
        \caption{BPIC'17.}
    \end{subfigure}

    \begin{subfigure}[t]{0.3\linewidth}
        \includegraphics[width=\linewidth]{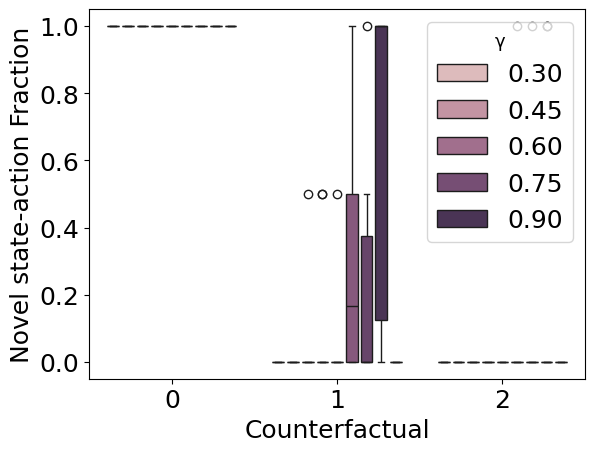}
        \caption{Greps.}
    \end{subfigure}
    \begin{subfigure}[t]{0.3\linewidth}
        \includegraphics[width=\linewidth]{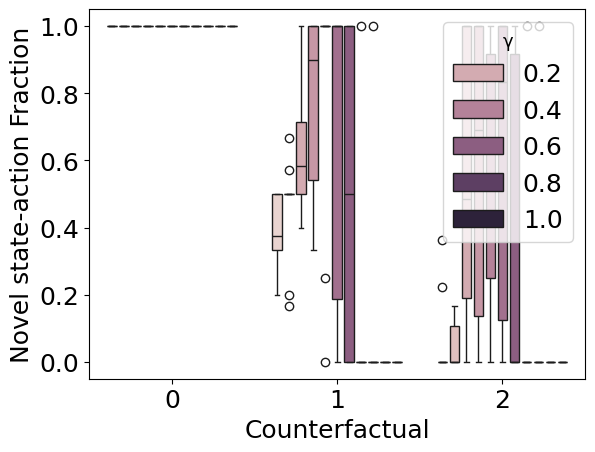}
        \caption{BPIC'12.}
    \end{subfigure}
    \begin{subfigure}[t]{0.3\linewidth}
        \includegraphics[width=\linewidth]{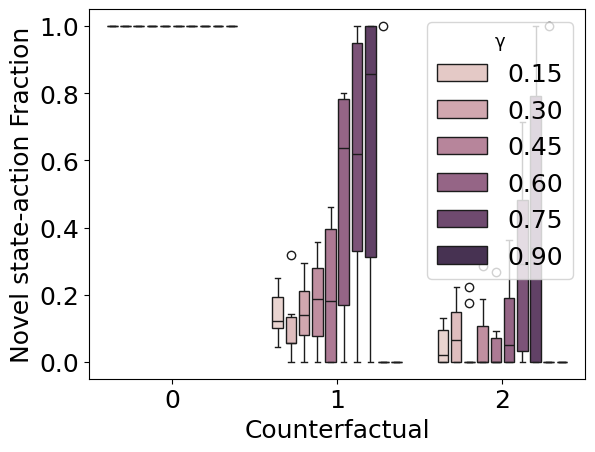}
        \caption{BPIC'17.}
    \end{subfigure}
    \caption{Results for diverse counterfactual strategies for each value of $\gamma$.}
    \label{fig:div_ce_gamma}
\end{figure*}


\begin{figure*}
    \centering
    \includegraphics[width=0.5\linewidth]{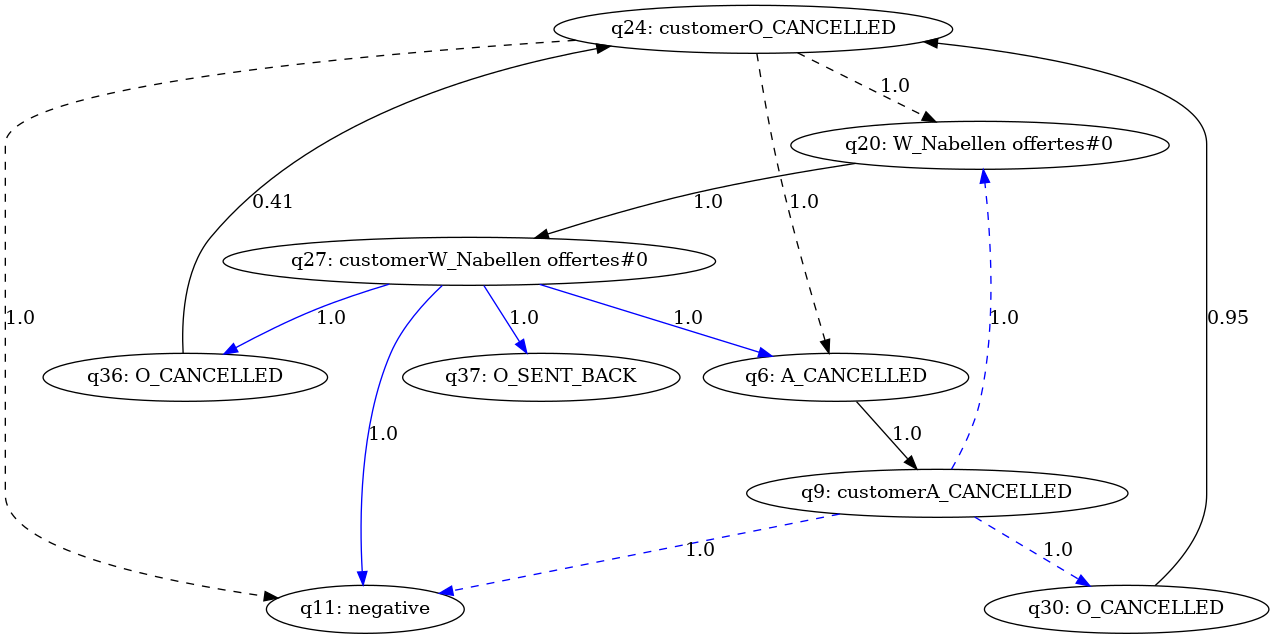}
    \caption{Comparison of two counterfactual strategies for BPIC'12.
      We only depict states that are close to a state where any
      counterfactual strategy differs from the initial strategy.
      Actions are colored blue if the first counterfactual strategy
      differs from the initial strategy for that action, and are
      dashed if the second counterfactual strategy differs.  }
    \label{fig:diversity_traversal}
\end{figure*}


\subsection{Further details on experimental setup}

We preprocess the datasets following the literature (for
GrepS~\cite{kobialkaSosym}, for
BPIC12~\cite{bautista2012bpic12challenge}, for
BPIC17~\cite{kobialka22edba,rodriguesStairwayValueMining}, and for
MSSD~\cite{johnsen2024nudging}). In the preprocessing of the datasets,
we removed irregular behavior, filtered unused events, and massaged
event names to improve the resulting models, e.g.\ we added the count
of the current offer in BPIC'12 to the event name to differentiate
consecutive offers.
For MSSD, We adapt the preprocessing by Johnsen~\etal for
counterfactuals by ignoring weights and real-time information, and
introduce a \emph{negative} state if the listening sessions have less
than 15 songs, and otherwise a \emph{positive} state.

Due to the complexity of MSSD, we generate models over multiples of $10\%$ of the $10\,000$ listening sessions
For each multiple of $10\%$, we build ten models each with a random subset from MSSD.
The generated models have between $64$ and $15\,000$ states, $76$ -- $33\,000$ transitions and a maximal out degree between $6$ and $5\,622$, as shown in Table~\ref{tab:model_overview}.
To differentiate between datasets and models, we denote models in bold, i.e.\ \textbf{MSSD10} denotes a model for 1000 listening sessions.
 
We implemented our evaluation framework in Python 3.10.12, using Gurobi 11.0.3\footnote{\url{https://www.gurobi.com/}}. 
For model construction, we use the jAlergia implementation of the automata learning algorithm Alergia~\cite{carrasco1994learning}, accessed through the Python automata learning library AAlpy, version 1.4.1~\cite{muvskardin2022aalpy}.
The generated benchmark set and code is available online.\footnote{\url{https://anonymous.4open.science/r/mdp_ce-12E2}}
We deploy the executions on
a workstation with 128 GB memory and a 64-core AMD EPYC processor.
We restrict Gurobi to use one core, limit its memory to \SI{2.5}{\giga\byte} of RAM, and the computation time to $1800$ seconds.

\subsection{Extended results}
We present extended results omitted in the main paper.
Table~\ref{tab:greps_bpic_quantiles} presents the run times for computing counterfactual strategies for each $\gamma$-quantile for \textbf{Greps}, \textbf{BPIC'12} and \textbf{BPIC'17}.

Table~\ref{tab:mssd_quantiles} presents categorical and run time results for each $\gamma$-sextile for computing counterfactual strategies for MSSD.


\end{document}